\pdfoutput=1

\documentclass[11pt]{article}

\usepackage[final]{acl}

\usepackage{times}
\usepackage{latexsym}
\usepackage{graphicx}
\usepackage{makecell}
\usepackage{multirow}
\usepackage{algorithm}
\usepackage{algorithmic}
\usepackage{amsfonts}
\usepackage{booktabs}
\usepackage{siunitx}
\usepackage{float}
\usepackage{enumitem} 
\usepackage{lipsum}
\usepackage{color}
\usepackage{booktabs}
\newcommand{\red}[1]{\textcolor{red!80}{#1}}

\usepackage{xcolor}
\usepackage{soul}
\usepackage{parskip}
\usepackage{colortbl}
\definecolor{lightBlue}{HTML}{A7C8DE}
\definecolor{DarkBlue}{HTML}{5790C0}
\definecolor{DarkGreen}{HTML}{3d6f28}
\definecolor{DarkRed}{HTML}{b02102}
\newcommand{\hlc}[2][yellow]{{%
    \colorlet{foo}{#1}%
    \sethlcolor{foo}\hl{#2}}%
}

\usepackage{pifont}
\usepackage{amsmath}
\usepackage{amssymb}
\usepackage{mathtools}
\usepackage{amsthm}
\usepackage[T1]{fontenc}

\usepackage[utf8]{inputenc}

\usepackage{microtype}

\usepackage{inconsolata}

    \title{\raisebox{-0.5cm}{\includegraphics[width=1.5cm]{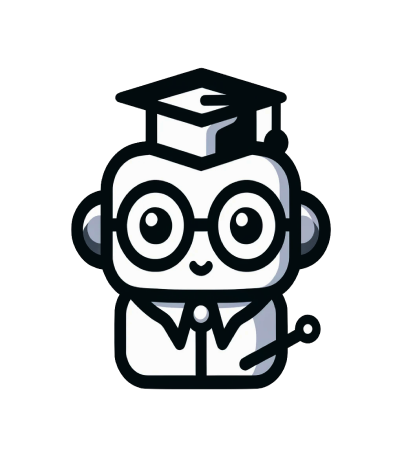}}FIRST: Teach A Reliable Large Language Model Through \\Efficient Trustworthy Distillation}

\DeclareSymbolFont{extraup}{U}{zavm}{m}{n}
\DeclareMathSymbol{\vardiamond}{\mathalpha}{extraup}{87}
\author{\bf Kashun Shum$^{\heartsuit*}$, ~ Minrui Xu$^{\heartsuit*}$, ~  Jianshu Zhang$^{\spadesuit*}$, ~ Zixin Chen$^{\heartsuit}$, ~ Shizhe Diao $^{\vardiamond}$\\ ~ \bf Hanze Dong$^{\heartsuit}$,  ~ Jipeng Zhang $^{\heartsuit}$, ~ Muhammad Omer Raza $^{\clubsuit}$\\
  $^{\heartsuit}$The Hong Kong University of Science and Technology, $^{\spadesuit}$Wuhan University\\
  $^{\vardiamond}$NVIDIA,   $^{\clubsuit}$ Purdue University \\
  \texttt{\{ksshumab, mxubh\}@connect.ust.hk}\\
  \\
}

\begin{document}
\maketitle
\def\thefootnote{*}\footnotetext{Equal Contribution.}
\def\thefootnote{\arabic{footnote}}

\begin{abstract}
Large language models (LLMs) have become increasingly prevalent in our daily lives, leading to an expectation for LLMs to be \textbf{trustworthy} --- both accurate and well-calibrated (the prediction confidence should align with its ground truth correctness likelihood). Nowadays, fine-tuning has become the most popular method for adapting a model to practical usage by significantly increasing accuracy on downstream tasks. Despite the great accuracy it achieves, we found fine-tuning is still far away from satisfactory trustworthiness due to "tuning-induced mis-calibration". In this paper, we delve deeply into why and how mis-calibration exists in fine-tuned models, and how distillation can alleviate the issue. Then we further propose a brand new method named E\textbf{F}f\textbf{I}cient T\textbf{R}ustworthy Di\textbf{ST}illation (\textbf{FIRST}), which utilizes a small portion of teacher's knowledge to obtain a reliable language model in a cost-efficient way. Specifically, we identify the "concentrated knowledge" phenomenon during distillation, which can significantly reduce the computational burden. Then we apply a "trustworthy maximization" process to optimize the utilization of this small portion of concentrated knowledge before transferring it to the student. Experimental results demonstrate the effectiveness of our method, where better accuracy (+2.3\%) and less mis-calibration (-10\%) are achieved on average across both in-domain and out-of-domain scenarios, indicating better trustworthiness.\footnote{The code is available at \url{https://github.com/SHUMKASHUN/FIRST}.}
\end{abstract}

\begin{figure}[!t]
    \begin{center}
    \includegraphics[width=\columnwidth]{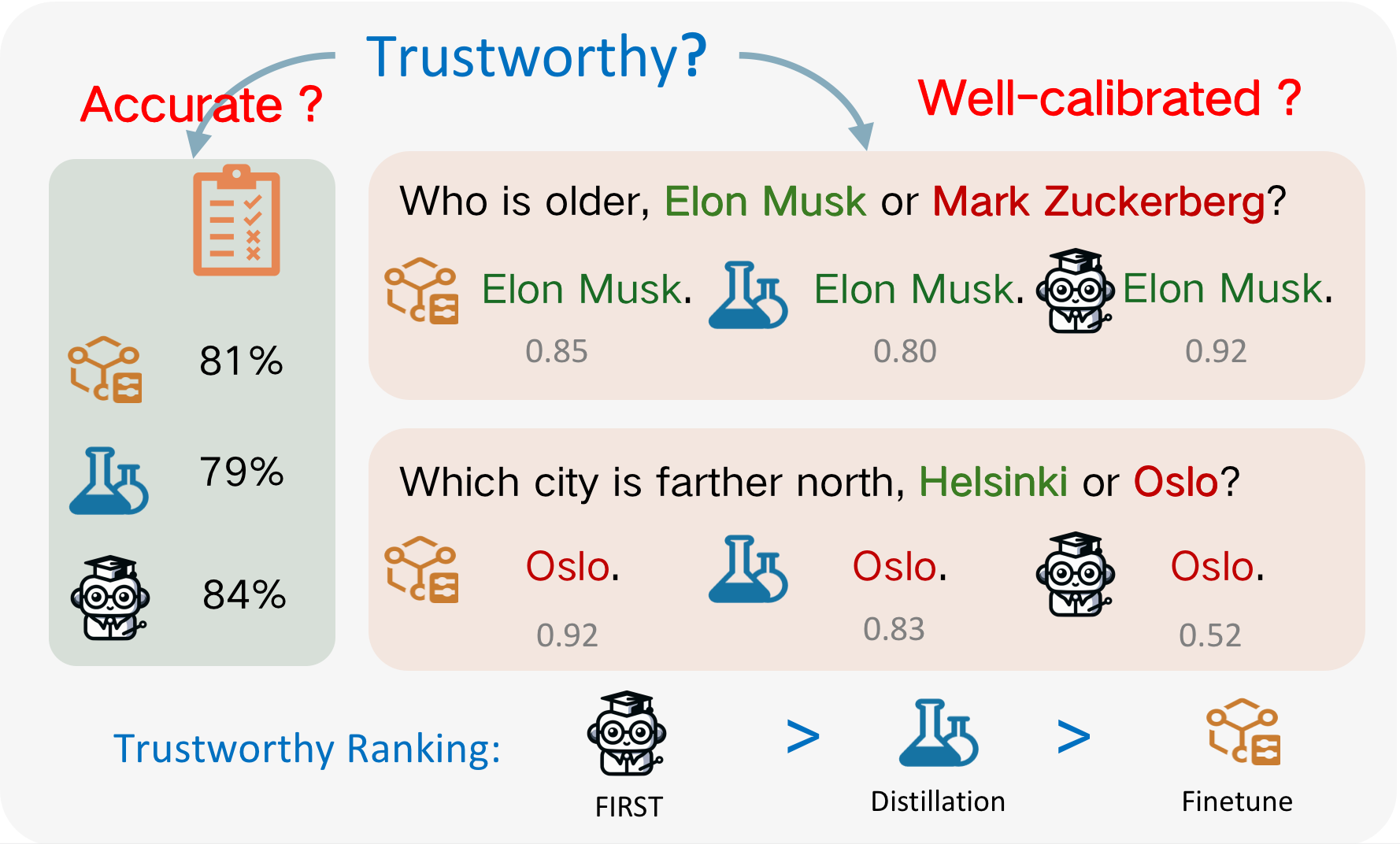}
    \vskip -0.1 in
    \caption{A trustworthy model should be both accurate (left) and well-calibrated (right). A well-calibrated model should produce high \textcolor{gray}{probabilities} for the \textbf{\textcolor{DarkGreen}{correct answer}} and low \textcolor{gray}{probabilities} for the \textbf{\textcolor{DarkRed}{wrong answer}}.}
    \label{fig:head}
    \end{center}
    \vskip -1 em
\end{figure}

\section{Introduction}
\label{introduction}

With the rapid development of large language models (LLMs), many powerful models have been deployed into our daily lives for practical usage to help us make decisions~\citep{yao2023react,sha2023languagempc,zhao2024expel}. This makes it urgent for us to know to what extent we can trust the outputs of the models. Calibration is one of the most important indicators beyond accuracy, which provides a confidence measure to the model’s predictions~\citep{guo2017calibration, hsieh-etal-2023-distilling}. In LLMs, confidence is exactly the probability for each generated token. Therefore, a well-calibrated model should align its prediction confidence with its ground-truth correctness likelihood as shown in Figure \ref{fig:head}. As an example, recent hallucination detection methods rely on model prediction confidence as a significant indicator of potential hallucination~\citep{zhang-etal-2023-enhancing-uncertainty,Varshney2023ASI}. If the model is incapable of giving accurate confidence levels, people may fail to detect hallucinations due to the model's over-confidence, or people may falsely identify hallucinations due to the model's under-confidence. Mis-calibration brings significant challenges for the deployment of LLMs in real-world applications.

Currently, there are two methods to obtain a language model for practical usage. First, fine-tuning, which fine-tunes pre-trained LLMs on specific datasets by matching each token entry with a target ground truth token. 
Although fine-tuning can consistently improve performance on downstream tasks~\citep{dodge2020finetuning, sun2020finetune, ziegler2020finetuning}, we identify that the model obtained in this way exhibits a nature of "tuning-induced mis-calibration".
Second, distillation-based methods transfer knowledge (e.g., soft labels) from larger LLMs to smaller models~\citep{gu2023knowledge}. Although distillation shows better calibration than fine-tuning as it matches each token entry with a probability distribution instead of a hard label, we find it is still biased because of the mis-calibration nature of teacher models. In addition, distillation faces the challenge of determining the optimal amount of knowledge to transfer. Transferring all the teacher's knowledge leads to high computational costs while transferring too little knowledge results in poor accuracy.
Therefore, it is crucial to balance between trustworthiness (accuracy and well-calibration) and efficiency for distillation-based methods.

To address the challenge of obtaining a trustworthy model, we propose e\textbf{F}f\textbf{I}cient t\textbf{R}ustworthy dis\textbf{T}illation (\textbf{FIRST}), aiming to efficiently utilize a relatively small amount of the teacher's knowledge. 
Specifically, we first identify the "concentrated knowledge" phenomenon, which shows that in the context of LLMs, the probability distribution of generated tokens is not uniform but rather concentrated on a few high-probability tokens. Based on this finding, we propose to use the top-5 tokens as the knowledge to balance the trade-off between storage space and the amount of knowledge transferred, achieving efficient distillation. 
Afterward, to eliminate the "tuning-induced mis-calibration" of the teacher model, we applied a "trustworthy maximization" to this portion of knowledge, ensuring that it maximizes the enhancement of the student model's accuracy while also guaranteeing its well-calibration. 

We first validate our method in in-domain scenarios, discovering that the models obtained by FIRST achieve excellent accuracy, even with the use of a relatively small amount of top-5 knowledge and the "trustworthy maximization" process can significantly enhance these models' robustness to mis-calibration. Furthermore, we test our approach in out-of-domain settings, demonstrating that models obtained by FIRST still exhibit the best trustworthiness and hold generalization ability. This indicates that FIRST enables smaller models to genuinely learn the capability of being trustworthy, rather than being confined to in-domain scenarios.

In summary, our key contributions include:

\begin{enumerate}[label=(\roman*)]
\item We discover that LLMs exhibit "concentrated knowledge" and "tuning-induced mis-calibration" phenomena, providing insights into obtaining trustworthy models.
\item We propose \textbf{FIRST}, which maximizes the effectiveness and trustworthiness of a relatively small portion of knowledge transferred from the teacher by "trustworthy maximization" to obtain a trustworthy student model. 
\item Extensive experiments demonstrate that models obtained using FIRST consistently achieve the highest level of trustworthiness across different settings.
\end{enumerate}


\section{Related Work}
\subsection{Trustworthy Models}
The current evaluation of LLMs predominantly focuses on accuracy, overlooking whether the models truly know the answer or are merely guessing (i.e. trustworthy). Recent works~\citep{sun2024trustllm,steyvers2024calibration} have demonstrated that accurate LLMs may not necessarily be "trustworthy" due to a significant calibration gap, so-called mis-calibration. This gap prevents us from trusting the output of the models, and it can further cause LLMs to generate harmful content, especially when subjected to adversarial attacks or jailbreak prompts~\citep{mo2024trustworthy,yao2024fuzzllm}. Our work further reveals how mis-calibration exists in different tuning methods and proposes a new trustworthy evaluation metric that covers both accuracy and calibration.

To achieve a well-calibrated LLM, recent work shows soft-label distillation shows better calibration ability~\citep{gu2023knowledge}. However, it still suffers from biased labels due to the mis-calibration nature of the fine-tuned teacher model. Our work is an improvement on this line of work by applying "concentrated knowledge" and "trustworthy maximization", leading to better accuracy, efficiency, and trustworthy.


\subsection{Knowledge Distillation}
Knowledge Distillation is a form of transfer learning that facilitates the transfer of knowledge from a larger teacher model to a smaller student model. The goal is to reduce the model size while maintaining or even improving performance. Based on whether we can access prediction probability, the existing distillation methods can be categorized into two types: Black-box Distillation and White-box Distillation.

Black-box Distillation refers to distillation from models that we are unable to access the weight and prediction logits such as PaLM~\citep{chowdhery2022palm}. Recent studies have attempted to distill reasoning ability from GPT~\citep{ho-etal-2023-large,shridhar-etal-2023-distilling} or some emergent ability such as chain-of-thought~\citep{hsieh-etal-2023-distilling,Li2023SymbolicCD}. However, these methods may still be categorized as the genre of data-augmentation-and-then-fine-tuning approaches.

\begin{figure}[!t]
    \begin{center}
    \includegraphics[width=0.95\columnwidth]{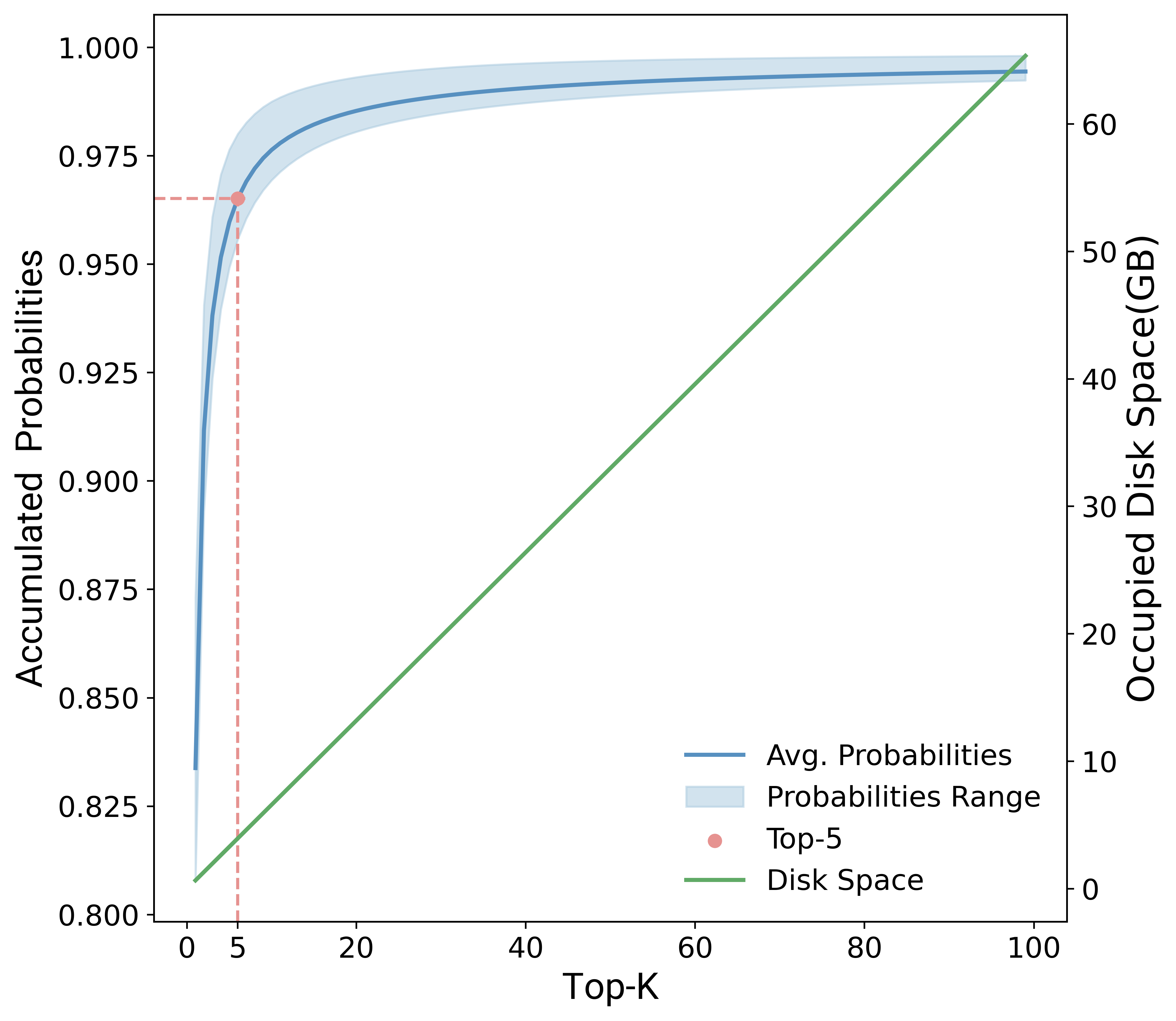}
    \vskip -0.1 in
    \caption{The blue line with range shows the averaged accumulated probability coverage for each token entry, from Top-1 to Top-100. \textbf{"Concentrated Knowledge"} : The red point represents accumulated probability for Top-5 tokens already exceed 95\%. The green line describes the disk usage if use Top-K token distribution during distillation.}
    \label{fig:why-top5}
    \end{center}
    \vskip -1 em
\end{figure}

White-box Distillation means the teacher models are either fully open-sourced such as Llama~\citep{touvron2023llama} or they can return partial probability distribution of the generated tokens, such as code-davinci-002. Instead of the hard token fine-tuning, white-box distillation typically uses more fine-grained signals by matching a distribution between teachers and students~\citep{gu2023knowledge,latif2023knowledge,agarwal2024onpolicy}. 
Further, in the field of white-box distillation, there are two different ways: online distillation and offline distillation. 
Online distillation~\citep{gu2023knowledge,zhou2023distillspec} needs to keep both the teacher model and the student model on the GPU simultaneously during training. 
On the other hand, offline distillation typically involves obtaining knowledge from the teacher model beforehand.
Our work is an extension of white-box offline distillation and focuses on how white-box offline distillation can be improved in terms of trustworthiness by re-calibrating the teacher distribution.


\label{exp:top5}
\begin{figure}[!t]
    \begin{center}
    \includegraphics[width=\columnwidth]{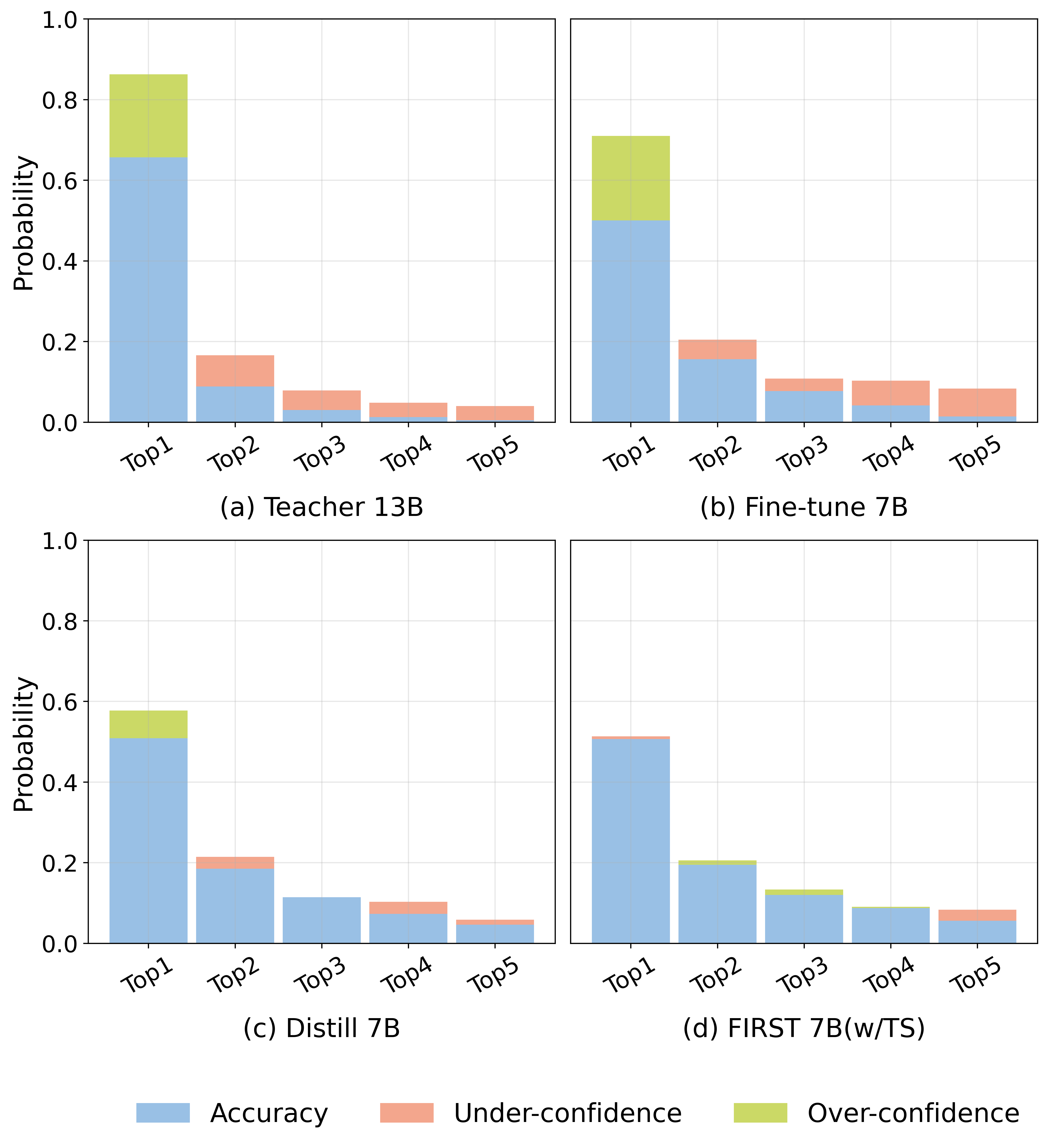}
    \vskip -0.1 in
    \caption{\textbf{"Tuning-induced Mis-calibration"} : Position-wise prediction probabilities with corresponding actual accuracy of (a) fine-tuned teacher model and (b) fine-tuned small model, (c) distilled model and (d) model produced by FIRST. }
    \label{fig:motivation}
    \end{center}
    \vskip -1 em
\end{figure}

\section{Preliminaries}
\subsection{Concentrated Knowledge}
\label{top5}
In the process of searching for a suitable trade-off between the amount of knowledge to transfer from the teacher model and efficiency, we begin by visualizing the probability distribution for each token entry. As illustrated in Figure~\ref{fig:why-top5}, the blue line with range describes how averaged accumulated probabilities increase when we select more tokens (ranked from highest probability to lowest probability in one entry). The trend clearly shows a few top-position tokens take most of the probability information of a token entry. To be specific, the accumulated probabilities of top-5 tokens can occupy over 95\% probabilities while the remaining 49995 (i.e. a model with vocab. size of 50k) tokens have nearly 0 probability. We named this phenomenon "Concentrated Knowledge" as almost full knowledge of a token entry is stored in its top-k tokens where the remaining tokens have negligible information.
\subsection{Tuning-induced Mis-calibration}
\label{Mis-calibration}
In the context of LLMs, mis-calibration can be divided into two types: over-confidence and under-confidence. Over-confidence occurs when the predicted probability of a token is higher than its actual accuracy, while under-confidence takes place when the predicted probability is lower than the actual accuracy. 

During the fine-tuning process of LLMs, cross-entropy loss is commonly employed, which encourages the models to assign a probability of 1 to one token and 0 to all other tokens based on the ground-truth token. This training nature results in 1.) an over-estimation of the ground truth token's probability and  
2.) an under-estimation of all other token's probability. As shown in Figure~\ref{fig:motivation} (a) and (b), it is observed that both fine-tuned LLMs exhibit over-confidence in their top-1 token predictions, while demonstrating under-confidence in the subsequent tokens. This phenomenon, which we call "tuning-induced calibration", highlights the untrustworthy nature of fine-tuned models.

Since fine-tuned teacher models suffer from this tuning-induced mis-calibration, if the knowledge from the mis-calibrated teacher models is directly used in traditional distillation-based methods, the student models are very likely to inherit the same mis-calibration nature as depicted in Figure~\ref{fig:motivation} (c). Motivated by the tuning-induced mis-calibration, our proposed method incorporates a "trustworthy maximization" procedure to re-calibrate the knowledge derived from the teacher models. This enables us to obtain a genuinely trustworthy student model.



\subsection{Expected Calibration Error}
To measure calibration in the context of LLMs, we adapt the expected calibration error (ECE) to the free-text generation task by treating the generation of a single token as a classification task. In this adaptation, we restrict the model to generate only one token from a set of candidate choices (e.g., A/B/C/D). For each token, we obtain the highest probability choice using $\mathop{\arg\max}_{i\in C} P(i)$, where $C$ represents the set of candidates. The probability of the chosen token is taken as the predicted confidence, and we calculate the accuracy by comparing the predicted choice to the ground truth. Then we utilize a total $M$ probability intervals as bins and categorize each chosen token into $m$-th bin according to the predicted confidence. The ECE~\citep{guo2017calibration} can be computed as follows:
\begin{equation}
\label{eq: ece}
    ECE = \sum_{m=1}^M \frac{|B_m|}{n} \lvert acc(B_m) - conf(B_m) \rvert
\end{equation}
Here, $M$ is the number of bins. $B_m$ represents the set of predictions in bin $m$, $|B_m|$ is the number of prediction instances in bin $m$, and $n$ is the total number of predictions. $acc(B_m)$ is the average accuracy of predictions in bin $m$, and $conf(B_m)$ is the average confidence of predictions in bin $m$. A lower ECE value indicates that the model’s predicted probabilities are more consistent with actual outcomes, meaning the model is better calibrated.

\subsection{Trustworthy Score}
When evaluating the trustworthiness of a model, it is essential to consider both high accuracy and effective calibration. Existing benchmarks primarily focus on accuracy, assuming that higher accuracy implies greater trustworthiness. However, our discovery of the widespread issue of "tuning-induced mis-calibration" has highlighted the inadequacy of relying solely on accuracy for a comprehensive evaluation of model trustworthiness. To address this limitation, we propose Trust Score metric to quantify a model's trustworthiness, which considers two key aspects: its ability to provide accurate answers (measured by $Acc$) and its capacity to align predicted confidences with actual accuracies (measured by $ECE$). The Trust Score is defined as follows:
\begin{equation}
\label{eq: trust}
    Trust = Acc - ECE
\end{equation}
By incorporating the Trust Score, we achieve a more balanced evaluation of trustworthiness, taking into account both accuracy and calibration.

\section{Efficient Trustworthy Distillation}
\begin{figure*}[t]
    \begin{center}
    \includegraphics[width=0.8\textwidth]{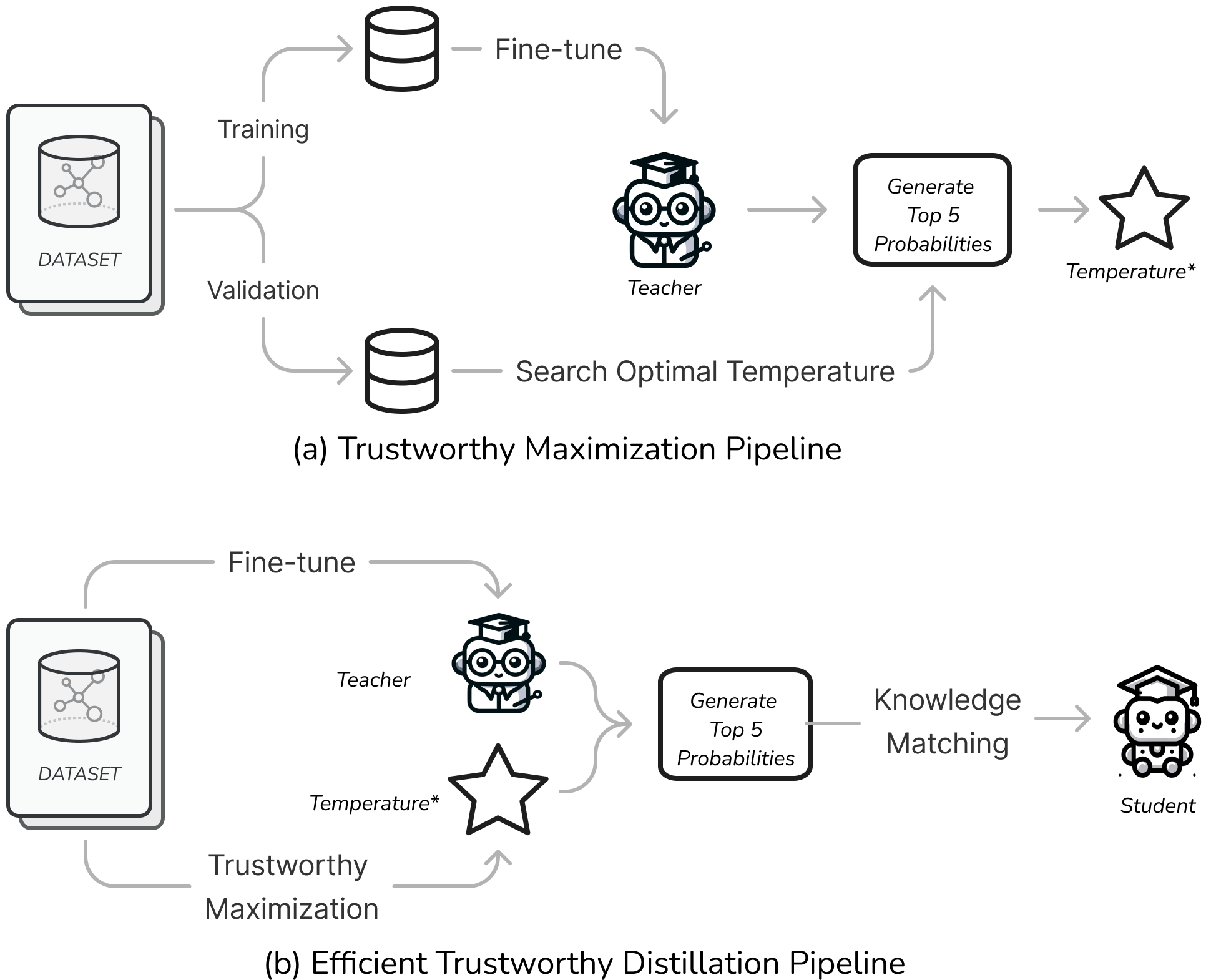}
    \vskip -0.5 em
    \caption{(a) The Trustworthy Maximization Step: we first fine-tune our the teacher model and then generate top-5 probabilities of all tokens and run a grid search to select the optimal temperature based on the validation set. (b) The overall Efficient Trustworthy Distillation Pipeline: based on the selected optimal temperature from (a), we obtain a well-calibrated student model by knowledge matching between student's knowledge and the portion of teacher knowledge.}
    \label{fig:pipeline}
    \end{center}
    \vskip -1 em
\end{figure*}
In this section, we introduce e\textbf{F}f\textbf{I}cient t\textbf{R}ustworthy dis\textbf{T}illation (\textbf{FIRST}), which can be divided into three parts. Firstly, we select top-5 tokens as knowledge for transfer (Efficient Knowledge Selection) in Sec. \S\ref{sec: selection}. Then, we adjust the knowledge for trustworthiness to ensure that the subsequent smaller models can maximize its utility (Knowledge Trustworthy Maximization) in Sec. \S\ref{sec: trustworthy max}. Finally, we describe the learning process of the student model (Knowledge Matching) in Sec. \S\ref{sec: distribution match}. The overall pipeline is shown in Figrue~\ref{fig:pipeline}.

\subsection{Efficient Knowledge Selection}
\label{sec: selection}
Transferring knowledge directly from teachers to students can be computationally costly and storage-intensive. For example, if we consider a vocabulary size of 50,000 tokens, retrieving the complete probability distribution from a dataset of 100,000 samples, with an average length of 2,048, would require a staggering 120 TB of storage, which is impractical.

Based on the discovery of "concentrated knowledge" in teacher LLMs, we observe that the majority of knowledge is concentrated within a small portion of top-position tokens, as elaborated in Section~\S\ref{top5}. Therefore, considering that both computation and disk space increase linearly with the number of selected token entries, we argue that it is not necessary to use the complete probability distribution. Instead, by selecting a small amount of top-position tokens that contain majority of knowledge, we can strike the optimal balance between computational overhead and effectiveness. As depicted in Figure~\ref{fig:why-top5}, accumulated probability of top-5 token entries occupy more than 95\% probabilities while reducing storage from 120 TB to 1.2 GB.



\subsection{Trustworthy Maximization}
\label{sec: trustworthy max}
Once the top-5 tokens and their corresponding probabilities are collected from the teacher model, it is crucial to subject this knowledge to further processing to ensure proper calibration, as teacher models can also suffer from "tuning-induced mis-calibration" due to fine-tuning (as we elaborate in Sec. \S\ref{Mis-calibration}). This additional calibration step ensures that the student model improves in both accuracy and trustworthiness.

\paragraph{Label Smoothing:} Similar to \citet{rafael2019labelsmooth}, we first attempted to address tuning-induced mis-calibration" by applying a smoothing coefficient, denoted as $\delta$, to mitigate the teacher model's over-confidence in its top-1 token predictions while alleviating under-confidence in other predicted tokens as follows:
\begin{equation}
\label{eq: ls}
\begin{cases}
P_T(i):=P_T(i)-\delta & \text{if }i=1\\
P_T(i):=P_T(i)+\frac{\delta}{4} & \text{if }2\leq i\leq 5
\end{cases}
\end{equation}
Here, $T$ denotes the teacher model, $P_T(i)$ represents the probability of the $i$-th top token. While label smoothing can effectively mitigate over-confidence in top-1 token predictions, we have identified significant drawbacks associated with this approach. Firstly, directly applying label smoothing may compromise the preservation of token rankings, particularly between the top-1 and top-2 tokens. This can lead to a decline in model performance in certain cases. Secondly, label smoothing uses a constant probability, disregarding the varying levels of over-confidence or under-confidence in different token entries. Consequently, this can result in a transition from under-confidence to over-confidence among the top 2-5 tokens, making it challenging to achieve a balanced calibration across all of them.

\paragraph{Temperature Scaling:} Subsequently, we explore another approach using a temperature scaling technique~\citep{guo2017calibration} to re-calibrate the probabilities:
\begin{equation}
\label{eq :ts}
    P_T(i) = \frac{\exp(P_T(i)/c)}{\sum_j \exp(P_T(j)/c)}
\end{equation}
This method offers several advantages. First, it allows for a more fine-grained adjustment of the probability distribution by controlling the temperature scaling parameter $c$, which can be optimized to achieve the lowest ECE values. Second, unlike label smoothing, temperature scaling can effectively balance the confidence levels of both top-1 and subsequent tokens, reducing both over-confidence and under-confidence issues by preserving token rankings and avoiding transition between under-confidence and over-confidence. 

This results in a more consistent and reliable calibration across all tokens, thereby enhancing the overall trustworthiness of the knowledge. Additionally, we find that selecting the optimal $c$ parameter on the validation set to maximize the knowledge can significantly enhance the effectiveness of transferring trustworthy knowledge. The knowledge processed by using this $c$ yields the best results for the student model (detailed in Sec. \S\ref{exp:c}). Due to the low cost of selecting $c$ on the validation set, we can tailor different $c$ values for different tasks. This demonstrates "temperature scaling" excellent scalability and flexibility.

\subsection{Knowledge Matching}
\label{sec: distribution match}
After obtaining the re-calibrated probability data $P_T$ that contains $ P_T(1),P_T(2),\dots,P_T(5)$, we use the same training data to train the student model. Instead of utilizing language modeling loss on hard labels, the probabilities of the 5 tokens that correspond to the teacher's top-5 of the student model are retrieved as $P_S$ which contains $ P_S(1), P_S(2),..., P_S(5)$. Kullback–Leibler divergence is then used to measure the loss between the teacher model and the student model:
\begin{equation}
    Loss(y_{1:N}) = \sum_{t=1}^N D_{KL}(P_T || P_S)
\end{equation}

\begin{table*}[!t]
\footnotesize
\centering
    \resizebox{\textwidth}{!}{
    \begin{tabular}{c|cccccc|cccccc}
    \toprule

     & \multicolumn{6}{c}{\textsc{\textbf{In-Domain}}} & \multicolumn{6}{c}{\textsc{\textbf{Out-of-Domain}}} \\
     
     &\multicolumn{3}{c}{CSQA}& \multicolumn{3}{c}{BoolQ} &\multicolumn{3}{c}{CSQA}& \multicolumn{3}{c}{OBQA}\\
     \cmidrule(lr){2-4} \cmidrule(lr){5-7} \cmidrule(lr){8-10} \cmidrule(lr){11-13}
     & $ECE \downarrow$ & $Acc \uparrow$ & $Trust \uparrow$ & $ECE \downarrow$ & $Acc \uparrow$ & $Trust \uparrow$ & $ECE \downarrow$ & $Acc \uparrow$ &  $Trust \uparrow$ & $ECE \downarrow$ & $Acc \uparrow$ & $Trust \uparrow$ \\
      \midrule
     & \multicolumn{12}{c}{\textsc{Llama 1 : 33B $\rightarrow$ 7B}}\\
     \midrule
     \textcolor{gray}{Teacher}\textsubscript{ 33B} & \textcolor{gray}{10.2} & \textcolor{gray}{82.4} & \textcolor{gray}{72.2} & \textcolor{gray}{7.7} & \textcolor{gray}{89.7} & \textcolor{gray}{82} & \textcolor{gray}{18.6} & \textcolor{gray}{69.2} & \textcolor{gray}{50.6} & \textcolor{gray}{20.2} & \textcolor{gray}{64.4} & \textcolor{gray}{44.2} \\
     Fine-tune\textsubscript{ 7B} & 11.8 & 79.9 & 68.1 & 6.5 & 82.5 & 76 & 12.5 & 48.2 & 35.7 & 21.9 & 43.4 & 21.5\\
     Distill\textsubscript{ 7B} & 9.4 & 78.9 & 69.5& \textbf{4.0} & 85.3 & 81.3 &  5.3 & 43.1 & 37.8 & 18.1 & 39.8 & 21.7\\
    Distill\textsubscript{ 7B w/ LS} & 9.1 & 78.1 & 69 & 19.0 & 85.3 & 66.3 & 5.2 & 43.9 & 38.7& 19.0 & 37.6 & 18.6 \\
     \rowcolor{lightBlue} FIRST\textsubscript{ 7B w/ TS} & \textbf{2.9} & \textbf{80.8} & \textbf{77.9} & \textbf{4.0} & \textbf{85.7} & \textbf{81.7} & \textbf{4.6} & \textbf{50.0} & \textbf{45.4} & \textbf{7.1} & \textbf{47.2} & \textbf{40.1}\\
     FIRST to Fine-tune & $\red{\uparrow_{8.9}}$ & $\red{\uparrow_{0.9}}$ & $\red{\uparrow_{9.8}}$ & $\red{\uparrow_{2.5}}$ & $\red{\uparrow_{3.2}}$ & $\red{\uparrow_{5.7}}$ & $\red{\uparrow_{7.9}}$ & $\red{\uparrow_{1.8}}$ & $\red{\uparrow_{8.7}}$ & $\red{\uparrow_{14.8}}$& $\red{\uparrow_{3.8}}$ & $\red{\uparrow_{18.6}}$\\
     \midrule
     & \multicolumn{12}{c}{\textsc{Llama 2 : 13B $\rightarrow$ 7B}}\\
     \midrule
     \textcolor{gray}{Teacher}\textsubscript{ 13B} & \textcolor{gray}{12.0} & \textcolor{gray}{81.6} & \textcolor{gray}{69.6} & \textcolor{gray}{6.8} & \textcolor{gray}{89.7} & \textcolor{gray}{82.9} & \textcolor{gray}{20.8} & \textcolor{gray}{65.7} & \textcolor{gray}{44.9} & \textcolor{gray}{28.7} & \textcolor{gray}{58.3} & \textcolor{gray}{29.9} \\
     Fine-tune\textsubscript{ 7B} & 14.0 & 76.8 & 62.8 & 8.4 & 87.5 & 79.1 & 21.2 & 50.0 & 28.8 & 30.1 & 45.6 & 15.5 \\
     Distill\textsubscript{ 7B} & 10.9 & 80.0 & 69.1 & 4.0 & 85.3 & 81.3 & 7.7 & 50.9 & 43.2 & 12.5 & 46.6 & 34.1 \\
     Distill\textsubscript{ 7B w/ LS} & 10.3 & \textbf{80.4} & 70.1 & 3.9 & 87.5 & 83.6 & 7.5 & 51.1 & 43.6 & 16.2 & 47.6 & 31.4 \\
     \rowcolor{lightBlue} FIRST\textsubscript{ 7B w/ TS} & \textbf{6.3} & 80.3 & \textbf{74} & \textbf{1.4} & \textbf{87.9} & \textbf{86.5} & \textbf{5.5} & \textbf{51.4} & \textbf{45.9} & \textbf{8.1} & \textbf{49.5} & \textbf{41.4} \\
     FIRST to Fine-tune & $\red{\uparrow_{7.7}}$ & $\red{\uparrow_{3.5}}$ & $\red{\uparrow_{11.2}}$ & $\red{\uparrow_{7}}$ & $\red{\uparrow_{0.4}}$ & $\red{\uparrow_{7.4}}$ & $\red{\uparrow_{15.7}}$ & $\red{\uparrow_{1.4}}$ & $\red{\uparrow_{17.1}}$ & $\red{\uparrow_{22}}$& $\red{\uparrow_{3.9}}$ & $\red{\uparrow_{25.9}}$\\
     \midrule
     & \multicolumn{12}{c}{\textsc{OpenLlama : 13B $\rightarrow$ 7B}}\\
     \midrule
     \textcolor{gray}{Teacher}\textsubscript{ 13B} & \textcolor{gray}{13.2} & \textcolor{gray}{78.5} & \textcolor{gray}{65.3} & \textcolor{gray}{7.5} & \textcolor{gray}{87.6} & \textcolor{gray}{80.1} & \textcolor{gray}{16.7} & \textcolor{gray}{49.5} & \textcolor{gray}{32.8} & \textcolor{gray}{13.4} & \textcolor{gray}{50.0} & \textcolor{gray}{36.6} \\
     Fine-tune\textsubscript{ 7B} & 10.5 & 75.0 & 64.5 & 3.6 & 81.5 & 77.9 & 21.6 & 28.3 & 6.7 & 16.1 & 30.4 & 14.3 \\
     Distill\textsubscript{ 7B} & 9.2 & 75.2 & 66 & 6.2 & 83.8 & 77.6 & 9.7 & 27.7 & 18 & 13.7 & 29.8 & 16.1 \\
     Distill\textsubscript{ 7B w/ LS} & 9.6 & 74.5 & 65.9 & 3.3 & 83.3 & 80 & 4.1 & 29.2 & 25.1 & 14.2 & 29.8 & 15.6 \\
     \rowcolor{lightBlue} FIRST\textsubscript{ 7B w/ TS} & \textbf{5.0} & \textbf{77.2} & \textbf{72.2} & \textbf{2.7} & \textbf{84.7} & \textbf{82} & \textbf{2.9} & \textbf{30.5} & \textbf{27.6} & \textbf{8.2} & \textbf{30.8} & \textbf{22.6} \\
     FIRST to Fine-tune & $\red{\uparrow_{5.5}}$ & $\red{\uparrow_{2.2}}$ & $\red{\uparrow_{7.7}}$ & $\red{\uparrow_{0.9}}$ & $\red{\uparrow_{3.2}}$ & $\red{\uparrow_{4.1}}$ & $\red{\uparrow_{18.7}}$ & $\red{\uparrow_{2.2}}$ & $\red{\uparrow_{20.9}}$ & $\red{\uparrow_{7.9}}$& $\red{\uparrow_{0.4}}$ & $\red{\uparrow_{8.3}}$\\

    \bottomrule
     
    \end{tabular}}
    \caption{Smaller models obtained by our method \colorbox{lightBlue}{FIRST} consistently achieves high accuracy $Acc$ across various scenarios while maintaining a low expected calibration error $ECE$ (see Eq. \ref{eq: ece}). The higher trust scores $Trust$ (see Eq. \ref{eq: trust}), the more trustworthy models are. Note that in the out-of-domain setting, we only obtain smaller models by fine-tuning or distilling on Alpaca, with CSQA and OBQA being unseen in this context, validating the generalizability of our approach. $\uparrow$ represents the larger the better while the $\downarrow$ means the smaller the better. \textbf{Bold} represents the best.}
    \label{tab:main_result}
    \vskip -0.1in

\end{table*}

\section{Experiment}
\subsection{Experimental Settings}
Our experiments focus on both In-Domain and Out-of-Domain settings to ensure generalization abilities. In the \textbf{In-Domain setting}, we utilize CommonsenseQA (CSQA) \citep{talmor2019commonsenseqa} and BoolQ \citep{clark2019boolq} for both training and testing. In the \textbf{Out-of-Domain setting}, we fine-tune and distill smaller models on a commonly used instruction-following dataset, Alpaca \citep{alpaca}, while, testing the models' performance over unseen task CommonsenseQA (CSQA) and OpenBook QA (OBQA) \citep{mihaylov-2018-suit}. This approach allows us to assess the generalization abilities of the smaller models on unseen tasks, simulating real-world scenarios where these models need to perform on unfamiliar tasks.

To ensure the practicality of our approach, we select three widely used model families for our experiments: Llama-1 \citep{touvron2023llama}, Llama-2 \citep{touvron2023llama2}, and OpenLlama \citep{openlm2023openllama}. In our experiments, we test four types of smaller models obtained through different methods: 

1) \textbf{Fine-tune\textsubscript{ 7B}}: Obtained by using fine-tuning with hard labels. 

2) \textbf{Distill\textsubscript{ 7B}}: Obtained by distillation methods without "knowledge trustworthy maximization". For a fair comparison with our approach, we also use the top-5 tokens as knowledge in the latter comparison.

3) \textbf{FIRST\textsubscript{ 7B w/TS}}: Obtained by our proposed method, primarily using temperature scaling (TS, see Eq. \ref{eq :ts}) within the trustworthy maximization phase.

4) \textbf{Distill\textsubscript{ 7B w/ LS}}: We also explore the use of label smoothing (LS, see Eq. \ref{eq: ls}) to show why we ultimately adopt TS over LS in "knowledge trustworthy maximization". In the latter experiments, we pick up the popular smoothing
coefficient 0.1 follow previous works~\citep{müller2020does}.

Additionally, we also provide the performance of \textbf{Teacher} models. For further implementation details, please refer to the Appendix~\ref{sec:detailed_experimental_setting}.



\subsection{Experiment Results}
\label{sec: main-exp}
Based on the results shown in Table \ref{tab:main_result}, we draw the following conclusions:

\noindent $\bullet$ \textbf{Fine-tuning lead to catastrophic mis-calibration}: We observed that although fine-tuned smaller models achieve relatively high accuracy in both in-domain and out-of-domain settings, their ECE values are notably high, resulting in overall low trust scores and lower reliability. 
This mis-calibration phenomenon is particularly pronounced in out-of-domain scenarios. For instance, we observe that the ECE of the model fine-tuned on OpenLllama 7B in the out-of-domain CSQA task reaches 21.6\%, while its accuracy is only 28.3\%, indicating that smaller models obtained through fine-tuning tend to be unreliable on tasks they have not been trained on.
In real-world scenarios, when smaller models are privately deployed, they will inevitably encounter tasks they have not been trained for. In such cases, there would be a mismatch between their confidence and true likelihood. They might confidently provide incorrect answers and even continuously emphasize their incorrect responses, thereby misleading users. This clearly does not meet the criteria of a trustworthy model.

\noindent $\bullet$ \textbf{Distillation brings bad calibration as well}:
Furthermore, distilled models without "Knowledge Trustworthy Maximization" show relatively bad calibration ability. For in-domain tasks, the distilled Llama-1 7B and Llama-2 7B have ECE values of 9.4\% and 10.9\% on CSQA, a mis-calibration level similar to fine-tuned models. And distilled model of OpenLlama shows even worse calibration than fine-tuned models on BoolQ. While for accuracy, it generally has an improvement over standard fine-tuning, but on some settings such as Llama-1 on CSQA, it also shows worse performance than fine-tuning. This suggests that direct distillation without further process the knowledge does not consistently lead to better calibration and performance.

\noindent $\bullet$ \textbf{Temperature Scaling outperforms Label Smoothing}: Here, we compare the results of different methods used in the  "Knowledge Trustworthy Maximization" phase. It is evident that FIRST\textsubscript{7B w/ TS} performs significantly better than Distill\textsubscript{7B w/ LS}. In the in-domain setting of BoolQ, the ECE values of FIRST\textsubscript{7B w/ LS} astonishingly reached 19.0\%, significantly worse than Distill\textsubscript{7B}, which does not apply any additional processing to the knowledge. This highlights that LS cannot deliver stable performance across all scenarios.
In contrast, FIRST\textsubscript{7B w/ TS} consistently achieves lower ECE in both in-domain and out-of-domain scenarios. Additionally, they attain better accuracy in most cases, resulting in the highest Trust Scores.

\begin{figure*}[t]
    \begin{center}
    \includegraphics[width=\textwidth]{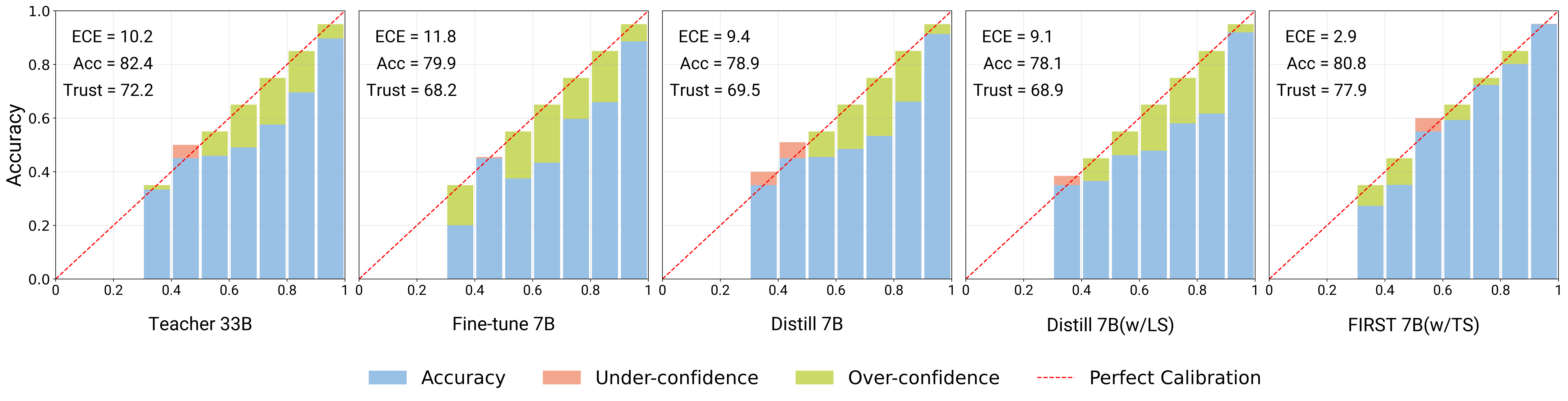}
    \vskip -0.5 em
    \caption{Reliability diagrams based on Llama-1 reveal the mis-calibration of various models on the CSQA dataset. In these diagrams, the X-axis is confidence divided into 10 bins, representing the model's confidence levels for each question's answer tokens. The Y-axis represents the accuracy within each bin. The red bar represents the degree to which the actual accuracy is higher than perfect calibration (under-confident), while the green bar means that the actual accuracy is lower than perfect calibration (over-confident).}
    \label{fig:visualize}
    \end{center}
    \vskip -1 em
\end{figure*}

\subsection{Reliability Analysis}

\paragraph{Reliability Diagrams.} To enhance our analysis and facilitate better comparisons, we employ reliability diagrams in addition to metric-based evaluations. As depicted in Figure~\ref{fig:visualize}, the reliability diagrams are divided into 10 bins based on the model’s confidence. The bars represent the expected accuracy within each bin, and the colors indicate whether the model is under-confident (red) or over-confident (green) within each bin. A perfectly calibrated model would have a straight diagonal line from the bottom left to the top right of such a diagram, indicating that the confidence level is exactly consistent with expected accuracy. 

The Fine-tune\textsubscript{7B} model exhibits catastrophic mis-calibration, primarily characterized by over-confidence in its predictions.
This means that the model tends to assign higher confidence levels to its predictions than what is justified by their actual accuracy. Although the Teacher\textsubscript{33B} model also suffers from over-confidence, its overall high accuracy results in a much higher trust score. Additionally, the Distill\textsubscript{7B} model demonstrates slightly improved calibration compared to the Fine-tune\textsubscript{7B} model. 
Remarkably, our FIRST\textsubscript{7B} model outperforms the other models, including the teacher model. It exhibits noticeably less under-confidence and over-confidence, as indicated by the smaller areas of the red and green bars, respectively, and its proximity to the perfect calibration line.

\subsection{Analysis of Top-5 Selection.}
\label{sec:analysis-of-top-5-selection}
Figure \ref{fig:why-top5} illustrates the disk space usage and cumulative probability coverage for knowledge selection ranging from the top-1 to the top-100 tokens. The blue line represents the average accumulated probabilities, while the shaded area indicates the range of probabilities. The green line shows the corresponding disk space required. The reasons we finally adopted top-5 are as follows:
\begin{enumerate}
    \item \textbf{Efficient Probability Coverage}:
    The figure demonstrates that selecting the top-5 tokens covers over 95\% of the total probability. This high coverage ensures that the majority of relevant knowledge is captured, making the distillation process effective.
    
    \item \textbf{Minimal Disk Space Usage}:
    The green line indicates the disk space required for storing the selected tokens. By selecting only the top-5 tokens, we significantly reduce the storage requirements compared to selecting more tokens. This efficiency is crucial for offline distillation, where disk space can be a limiting factor.
    
    \item \textbf{Balancing Trade-offs}:
    The top-5 selection strikes a balance between maximizing probability coverage and minimizing disk space usage. This balance ensures that the distilled knowledge is both comprehensive and storage-efficient, enabling practical implementation in various scenarios.
    
    \item \textbf{Scalability}:
    Our method exhibits strong scalability. It is naturally extendable to distillation from models such as the GPT-3 series (text-davinci-003), which can only return top-5 token probabilities. This increases the range of LLMs that can be used as teacher models, allowing student models to be effectively trained even in semi-black box scenarios. 
\end{enumerate}

\begin{figure}[!t]

    \begin{center}
    \includegraphics[width=\columnwidth]{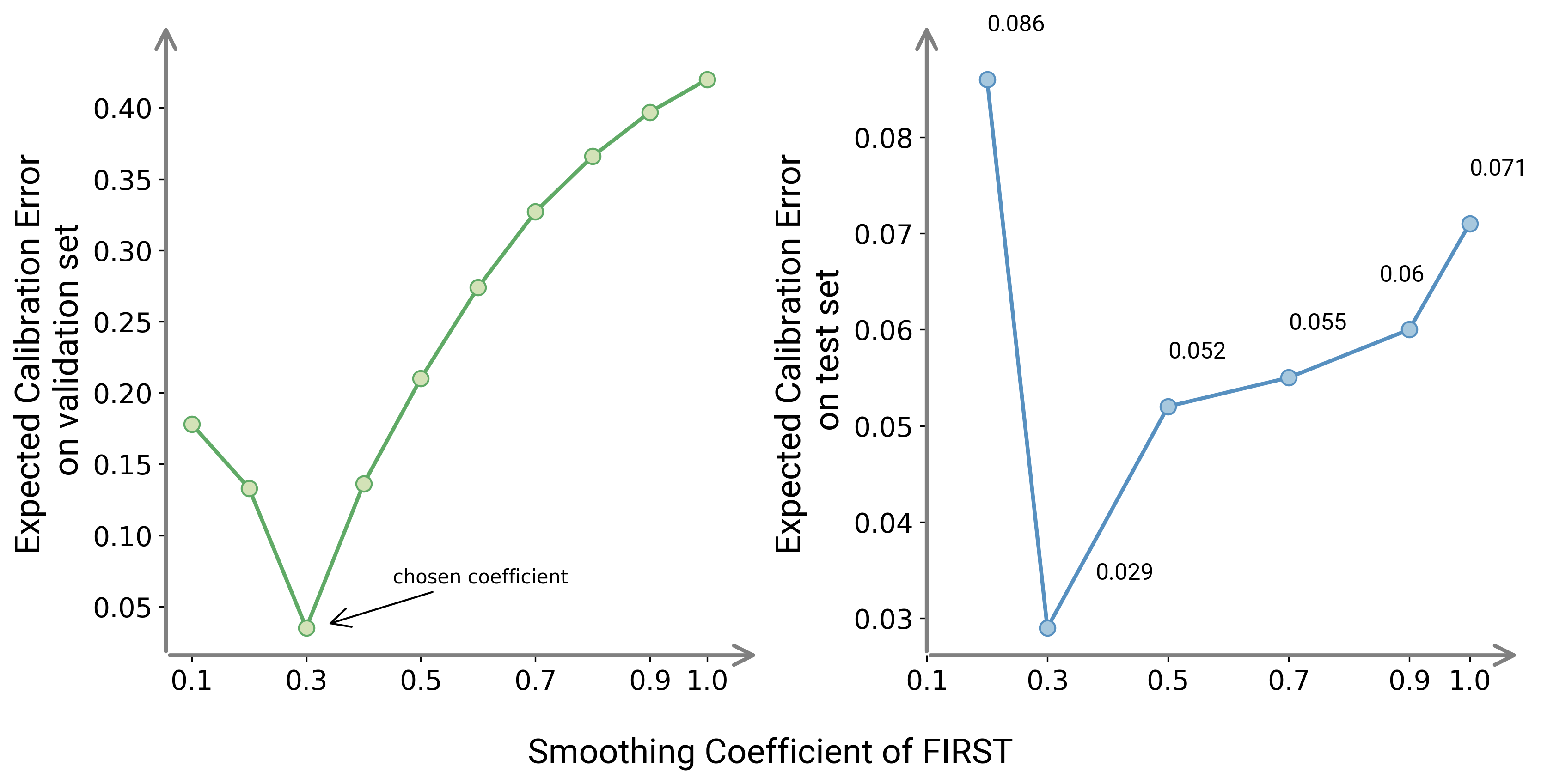}
    \vskip -0.1 in
    \caption{Left shows the comparison of different smoothing coefficients on the validation set, while the right part demonstrates its corresponding calibration effect on the test set.}
    \label{fig:coefficient}
    \end{center}
    \vskip -1 em
\end{figure}
\label{sec:additional_exp}

\subsection{Temperature Scaling Parameter Analysis}
\label{exp:c}
\noindent As described in the section on Knowledge Trustworthy Maximization (Sec. \S\ref{sec: trustworthy max}), we employ a temperature scaling parameter to optimize the ECE (Expected Calibration Error) value on the validation set, as illustrated in the left part of Figure~\ref{fig:coefficient}. By employing grid search, we initially partition the range from 0 to 1 into increments of 0.1 and identify the temperature associated with the lowest ECE value, for instance, 0.3. A larger temperature results in all top-5 tokens converging to the same probabilities, specifically 0.2 when the number of candidate choices is 5. When the temperature is set to 1, the probability of the top-1 token is dramatically compressed, while the probabilities of the other tokens are enlarged accordingly. Conversely, a temperature of 0.1 can even amplify the probabilities of over-confident tokens, leading to even worse calibration.

To further refine the search for the optimal temperature, we narrow down the interval and use a smaller step size of 0.02. This allows us to pinpoint the best temperature more precisely.
Additionally, we compare the performance of FIRST using the selected optimal temperature with other different temperatures as shown in the right part of Figure~\ref{fig:coefficient}. FIRST with optimal temperature do outperform those with other levels of temperatures with a large margin, indicating the effectiveness of selecting such optimal temperature.

\section{Conclusion}
In conclusion, our proposed method, e\textbf{F}f\textbf{I}cient t\textbf{R}ustworthy di\textbf{ST}illation (FIRST), effectively enhances both accuracy and calibration in large language models. By applying "trustworthy maximization", FIRST efficiently transfers the minimal yet most effective knowledge from teacher to student models. Experimental results show that FIRST consistently improves trustworthiness across various scenarios, demonstrating its potential to create reliable language models for practical applications.


\section{Limitations}
It is shown that our efficient trustworthy distillation (FIRST) demonstrates superior calibration ability and performance over direct distillation and standard fine-tuning methods.
However, despite these exciting results, there are still some limitations in our current work, as well as potential opportunities for future research.
\paragraph{Extend to Large Teacher Model}: Due to the resource limitation, our largest teacher model is Llama 33B, which is not very large but already achieving exciting results by distillation to a 7B student model. We expect that employing a large teacher model such as 70B can lead to better calibration ability and performance since a larger model learns a better distribution. However, we are unable to explore how very large teachers perform due to resource limitations.
\paragraph{Top-K Chosen in Offline Distillation:}Another limitation of this work is that it does not provide a rigorous study on how many token probabilities to choose for one entry is optimal for knowledge distillation in large language models. Currently, we consistently choose the top-5 token probability to retrieve because of the reasons stated in ~\S\ref{sec:analysis-of-top-5-selection}. However, how much token probability to use is optimal could be an important area for further exploration and development.


\bibliography{custom}
\appendix

\onecolumn
\begin{table*}[!t]
\scriptsize
\centering

    \begin{tabular}{l|c|c|c}
    \toprule

     & \textsc{Standard Fine-tuning} & \textsc{Direct Distillation} & \textsc{FIRST}\\
     \midrule
     Question & \multicolumn{3}{c}{Which city is farther north, Oslo or Helsinki?}\\
     \midrule
     Correct Answer & \multicolumn{3}{c}{Helsinki} \\
     \midrule
     Generated & 
     \hlc[DarkBlue]{Oslo} is farther north than Helsinki. &
     \multicolumn{1}{l|}{\hlc[DarkBlue]{Oslo} is farther north than Helsinki.} & 
     \multicolumn{1}{l}{\hlc[lightBlue]{Oslo} is farther north than Helsinki.} \\ 
     
     Confidence & \multicolumn{1}{l}{ \textcolor{DarkBlue}{0.92}$\;\rightarrow\;$\textcolor{DarkBlue}{over-confident} } & \multicolumn{1}{l}{ \textcolor{DarkBlue}{  0.83}$\;\rightarrow\;$\textcolor{DarkBlue}{over-confident}} & \multicolumn{1}{l}{\textcolor{lightBlue}{0.52}} \\
    \midrule
     \midrule
     Question & \multicolumn{3}{c}{Is Donald Trump a Neo-con American politician and businessman for the Republicans, with a long and varied career?}\\
     \midrule
     Correct Answer & \multicolumn{3}{c}{No} \\
     \midrule
     Generated & \multicolumn{1}{l|}{\hlc[DarkBlue]{Yes}.} &\multicolumn{1}{l|}{\hlc[DarkBlue]{Yes}.} & \multicolumn{1}{l}{\hlc[lightBlue]{Yes}.} \\ 

     Confidence & \multicolumn{1}{l}{ \textcolor{DarkBlue}{0.91}$\;\rightarrow\;$\textcolor{DarkBlue}{over-confident} } & \multicolumn{1}{l}{ \textcolor{DarkBlue}{  0.85}$\;\rightarrow\;$\textcolor{DarkBlue}{over-confident}} & \multicolumn{1}{l}{\textcolor{lightBlue}{0.54}} \\
     
     \midrule
     \midrule
     Question & \multicolumn{3}{c}{If I want to visit Beijing in spring, when should I go? Answer Choices: (a) June (b) July (c) August (d) September (e) October}\\
     \midrule
     Correct Answer & \multicolumn{3}{c}{None} \\
     \midrule
     Generated & \multicolumn{1}{l|}{\hlc[DarkBlue]{(c)}.} &\multicolumn{1}{l|}{\hlc[DarkBlue]{(d)}.} & \multicolumn{1}{l}{\hlc[lightBlue]{(d)}.} \\ 

     Confidence & \multicolumn{1}{l}{ \textcolor{DarkBlue}{0.58}$\;\rightarrow\;$\textcolor{DarkBlue}{over-confident} } &\multicolumn{1}{l}{ \textcolor{DarkBlue}{  0.41}$\;\rightarrow\;$\textcolor{DarkBlue}{over-confident}} & \multicolumn{1}{l}{\textcolor{lightBlue}{0.27}} \\
     
    \midrule
    \bottomrule

    \end{tabular}
    \caption{A case study on how fine-tuned model and direct distilled model tend to over-confident on the wrong answer with high confidence. While FIRST though outputs a wrong answer, it produces low confidence to show its uncertainty.}
    \label{tab:case}
\end{table*}

\section{Detailed Experimental Setting}
\label{sec:detailed_experimental_setting}

\subsection{Implementation Details}
\label{sec:implemenatation_details}

We train our models on 8 GPU (RTX A6000 48G) using the Adam optimizer with beta set to be [0.9, 0.999] and epsilon fixed to be 1e-6 and cosine annealing scheduler with a warm-up ratio of 0.03. For fine-tuning, we utilize LMFlow~\citep{diao2023lmflow} package to obtain a well fine-tuned model by a standard 3-epoch training and control the batch size to be 32 on each GPU and the learning rate for teacher models to be 2e-5. Finally, for distillation, the batch size is set to 32 on each GPU and we train our model for 3 epochs, the last checkpoint is used for evaluation since it has the best performance.

In addition, when implementing distillation without re-calibration, we use the following normalization function to normalize the top 5 distribution and prevent the probability to be 0.
$$
P_T(i)=\frac{P_T(j)+\delta}{\sum_j(P_T(j)+\delta)}
$$
In our setting, i, j = 1, . . . , 5, representing the top-5 token probability and $\delta$ is a small shift amount that prevent the probability to be 0 after normalization. The $\delta$ is set to be 1e-6 to minimize the influence.

\subsection{Prompt and Data Format}
\label{sec:prompt_format}
\begin{table*}[!t]
\small
\centering

\begin{tabular}{p{0.9\textwidth}}
\toprule
\underline{\textbf{CSQA}}\\
\textbf{Q: }The sanctions against the school were a punishing blow, and they seemed to what the efforts the school had made to change?\\
Answer Choices:\\
(a) ignore\\
(b) enforce\\
(c) authoritarian\\
(d) yell at\\
(e) avoid\\
\textbf{A: }The answer is (a).\\
\\
\underline{\textbf{OBQA}}\\
\textbf{Q: }food is a source of energy for what?\\
Answer Choices:\\
(A) waterfalls \\
(B) fires \\
(C) grass snakes \\
(D) mountains \\
\textbf{A: }The answer is (C).\\
\\
\underline{\textbf{Alpaca}}\\
Below is an instruction that describes a task, paired with an input that provides further context. Write a response that appropriately completes the request.\\
\\
\#\#\# Instruction:\\
For the given list of items, classify them into two categories.\\
\\
\#\#\# Input:\\
Carrot, Apple, Pumpkin, Orange\\
\\
\#\#\# Response:\\
Fruits: Apple, Orange\\
Vegetables: Carrot, Pumpkin\\
\\
\underline{\textbf{BoolQ}}\\
Below is an instruction that describes a task. Write a response that appropriately completes the request.\\
\\
\#\#\# Instruction:\\
Read the input passage and answer the question: is windows movie maker part of windows essentials? Your answer should be Yes or No.\\
\\
\#\#\# Input:\\
Windows Movie Maker (formerly known as Windows Live Movie Maker in Windows 7) is a discontinued video editing software by Microsoft. It is a part of Windows Essentials software suite and offers the ability to create and edit videos as well as to publish them on OneDrive, Facebook, Vimeo, YouTube, and Flickr.\\
\\
\#\#\# Response:\\
Yes\\
\bottomrule
\end{tabular}
\caption{Examples of our prompts and data formats for four different datasets. The same formats are used across all models and experiments.} 
\label{tab:prompt_dataset}
\end{table*}
For question-answering tasks, we follow ~\citet{shum2023automatic}'s format and fine-tune the model in a zero-shot setting. For out-of-domain tasks, we directly follow Alpaca's~\citep{alpaca} setting to obtain the fine-tuned model. The full prompt formats are shown in Table~\ref{tab:prompt_dataset}.

\section{Additional Analysis}
\subsection{Case Study}
We further conduct three case studies to show that FIRST indeed helps mitigate mis-calibration in real-world question answering. 

As shown in Table~\ref{tab:case}, we ask the models of three different tuning methods on Alpaca to answer the question:  \texttt{which city is farther north, Oslo or Helsinki?} The correct answer is \texttt{Helsinki} and the wrong answer is \texttt{Oslo}. From the output confidence, we can see that standard fine-tuned models and direct distillation give high confidence in the wrong answer, which is far from satisfactory for trustworthy in real-world settings, especially when additional post-processing procedures were expected to be applied to filter wrong answers by identifying unconfident responses. In comparison, FIRST greatly mitigates this mis-calibration by producing a confidence of around 50\% which indicates the model is not sure about the generated answer, allowing systems to filter those undesirable answers by a hard confidence threshold.

In the third case, we follow the FalseQA~\citep{hu2023wont}. In this case, all of the answer choices are expected to be wrong and models should output a confidence of 25\% in the top-1 token to achieve minimal ECE value. That's why our FIRST shows best calibration in this case.

\section{Trust Score Design}
Expected calibration error (ECE) is calculated by weighted average of difference between confidence and accuracy, which means accuracy and ECE are naturally in the same scale. Given that higher accuracy while lower ECE is better, it is intuitive and reasonable to define the trustworthy score by subtracting ECE from the accuracy. Besides, the product of ACC and ECE (e.g. $Trust \; Score = ACC \cdot (1 - ECE)$ ) will introduce a factor of ACC to the ECE score :
$ACC - ACC\cdot ECE$. This would bring unfairness when comparing large models (high accuracy) with small models (low accuracy) because we expect ECE has the same importance when evaluating either high-accuracy model or low-accuracy model.
\end{document}